\documentclass[sigconf, anonymous=false]{acmart}

\AtBeginDocument{%
  \providecommand\BibTeX{{%
    \normalfont B\kern-0.5em{\scshape i\kern-0.25em b}\kern-0.8em\TeX}}}

\setcopyright{acmcopyright}
\copyrightyear{2018}
\acmYear{2018}
\acmDOI{XXXXXXX.XXXXXXX}

\acmConference[Conference acronym 'XX]{Make sure to enter the correct
  conference title from your rights confirmation emai}{June 03--05,
  2018}{Woodstock, NY}
\acmPrice{15.00}
\acmISBN{978-1-4503-XXXX-X/18/06}

\usepackage{enumitem}
\usepackage{soul}
\usepackage{url}
\usepackage{hyperref}
\usepackage{pifont}
\usepackage{subcaption}  
\usepackage{multicol}
\usepackage{multirow}
\usepackage{threeparttable}
\usepackage{booktabs}
\usepackage{array}

\usepackage{xcolor}
\usepackage{colortbl}
\usepackage{caption}
\usepackage{amsmath}
\usepackage[outline]{contour}
\usepackage{makecell}
\usepackage{pifont}
\begin{document}

\title{VFed-SSD: Towards Practical Vertical Federated Advertising}



\author{Wenjie Li{\normalsize$^{1,2}$},
Qiaolin Xia{\normalsize$^{2}$},
Junfeng Deng{\normalsize$^{2}$},
Hao Cheng{\normalsize$^{2}$},
Jiangming Liu{\normalsize$^{2}$},
\\Kouying Xue{\normalsize$^{2}$},
Yong Cheng{\normalsize$^{2}$},
Shu-Tao Xia{\normalsize$^{1}$}}
\affiliation{\country{$^1$Tsinghua University,
$^2$Tencent}}
\email{liwj20@mails.tsinghua.edu.cn,
xiast@sz.tsinghua.edu.cn}
\email{{jolinxia, junfengdeng, dennycheng, jiangmliu,  lornaxue, peterycheng}@tencent.com}

\renewcommand{\shortauthors}{Trovato and Tobin, et al.}

\newcommand{\cmark}{\ding{51}}%
\newcommand{\xmark}{\ding{55}}%
\newcommand{\short}{\text{VFed-SSD }}
\newcommand{\ams}{\text{Game-Ad }}
\newcommand{\ola}{\text{Game-Social }}
\newcommand{\criteo}{\text{Criteo-NC }}
\newcommand{\myuparrow}{\contourlength{0.02em}\contour{black}{$\uparrow$}}
\newcommand{\myXm}{{\color{black}\xmark}}
\newcommand{\myCm}{{\color{black}\cmark}}
\newcommand{\XQL}[1]{\textcolor{blue}{#1}}
\newcommand{\myColor}[1]{{\color[HTML]{3166ff} #1}}

\begin{abstract}
As an emerging secure learning paradigm in leveraging cross-silo private data, vertical federated learning (VFL) is expected to improve advertising models by enabling the joint learning of complementary user attributes privately owned by the advertiser and the publisher. However, there are two key challenges in applying it to advertising systems: \textit{\textbf{a)} the limited scale of labeled overlapping samples}, and \textit{\textbf{b)} the high cost of real-time cross-silo serving}. In this paper, we propose a semi-supervised split distillation framework \textbf{\short} to alleviate the two limitations. We identify that: \textit{\textbf{i)} there are massive unlabeled overlapped data available in advertising systems}, and \textit{\textbf{ii)} we can keep a balance between model performance and inference cost by splitting up the federated model.} Specifically, we develop a self-supervised task \textit{\textbf{M}atched \textbf{P}air \textbf{D}etection} (\textbf{MPD}) to exploit the vertically partitioned unlabeled data and propose the \textit{\textbf{Split} \textbf{K}nowledge \textbf{D}istillation} (\textbf{SplitKD}) schema to avoid cross-silo serving. 
Empirical studies on three industrial datasets exhibit the effectiveness of our methods, with the median AUC over all datasets improved by \textbf{$\mathbf{0.86\%}$} and \textbf{$\mathbf{2.6}\%$} in the local and the federated deployment mode respectively. Overall, our framework provides an efficient solution for cross-silo real-time advertising with minimal deploying cost and significant performance lift.
\end{abstract}



\maketitle

\section{Introduction}
\begin{figure}[t]
\centering
\begin{subfigure}[b]{4cm}
         \centering
         \includegraphics[width=4cm]{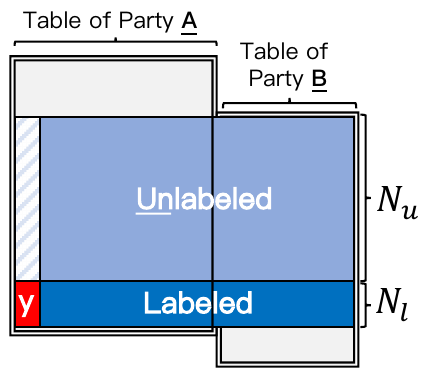}
         \caption{\small{The overlapped labeled samples are much fewer than unlabeled ones ($N_l << N_u$).}}\label{fig:motif1}
\end{subfigure}
\hfill
\begin{subfigure}[b]{4cm}
         \centering
         \includegraphics[width=4cm]{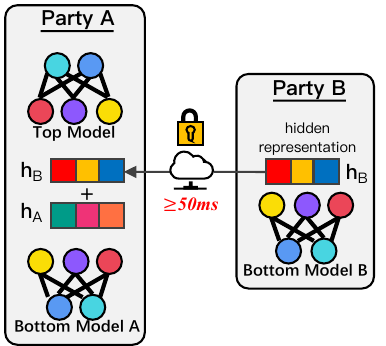}
         \small\caption{\small{Cross-party serving involves extra cost of encryption, network and overall system design.}}\label{fig:motif2}
\end{subfigure}
\caption{Two challenges of VFL in advertising systems.} \label{fig:limit}
\end{figure}

As two key agencies in advertising systems, the advertiser and the publisher (i.e. advertising platforms) own complementary data of user actions. Collectively leveraging these multi-platform user behaviors is helpful to achieve more fine-grained and comprehensive user interest modeling, which is beneficial to many tasks such as click-through rate (CTR) estimation and conversion rate (CVR) prediction. However, it is nearly impossible to directly share raw data across platforms due to data privacy regulations \cite{gdpr} and commercial confidentiality among service agencies. 

To address these issues, vertical federated learning \cite{vfl_survey1,vfl_survey2} has been explored to jointly utilize cross-platform attributes without compromising user privacy. In a typical VFL process, participants first execute a private set intersection (PSI\cite{peterson2019private}) to obtain the aligned (i.e., overlapped) dataset and then jointly train a distributed splitNN two-tower model \cite{splitNNVP,fldssm,blindfl}. However, the requirement of data overlapping and the distributed form of SplitNN
makes it impractical in industry applications (as shown in Figure \ref{fig:limit}): 
\begin{itemize}[leftmargin=*]
    \item \textbf{Limited training data}:
    Overlapped users between different business parties are usually limited and constitute only a small portion of the user population. Such reduction of available training data increases the risk of overfitting and result in low-quality embeddings and hidden representations, especially in advertising tasks with sparse and high-dimensional categorical features\cite{pan2019warm}.

    \item \textbf{Impractical cross-party inference}:
    The inference process of a splitNN federated model involves multiple participants differing in computation efficiency, IO speed and network condition. Such challenging cases make it harder to achieve reliable and real-time inference for million-wise QPS traffics which is typical in real-world ad systems\cite{scale_rtb_icdm,survey_rtb}.

\end{itemize}

In this paper, we propose the \textbf{S}emi-supervised \textbf{S}plit \textbf{D}istillation framework for VFL (\textbf{\short}) to tackle these issues. Our main motivations are:

\begin{itemize}[leftmargin=*]
    \item \textbf{Utilizing unlabeled data}:
    Due to temporal shifting of user interests, advertising models are frequently updated at a weekly or daily cadence using only latest fresh data \cite{muhamed2022dcaf}. Thus, there are lots of outdated historical data available in advertising system. Although the label (e.g., click and conversion actions) is almost the most volatile and usually noisy \cite{ltr_noise}, the attributes of these samples are more stable to temporal shifts and contains valuable information\cite{muhamed2022dcaf}. We argue that by including these historical data in PSI, we can get massive overlapped and unlabeled samples. Therefore, we develop a self-supervised task MPD to exploit these unlabeled data. Specifically, we use MPD to learn a pre-training model and use it to warm-up down-stream task models.

    \item \textbf{Distilling federated models}: To minimize system requirements of applying VFL in online serving, We develop split knowledge distillation (SplitKD) to distilling federated models as local models. The student model can be deployed independently for inference, thus free from building a cross-silo inference system. We show that (Figure \ref{tab:vkd_result}), such incomplete distillation schema is capable to maintain advantage over local models and can be further enhanced by MPD pre-training.
\end{itemize}

\noindent Our contributions can be summarized as follows:
\begin{enumerate}[leftmargin=*]
    \item We develop a simple and effective self-supervised task MPD for VFL to leverage unlabeled data. It is easy to implement in industry application and well-motivated by the theoretical principle of maximizing point-wise mutual information.
    \item We design a federated distillation method SplitKD to acquire federation-aware local models. It is easy to deploy in real-world ad systems and more effective than local models.
    \item Our framework is validated on both industry and public datasets, by achieving significant AUC improvement (\textbf{$\mathbf{0.5\sim1.2}$} in percentile) compared to vanilla local models.
\end{enumerate}

\section{Related Work}\label{sec:rl}
\noindent \textbf{Federated Learning}. Our paper focus on improving the utility and efficiency of recommendation tasks in two-party VFL settings. This motivation clearly distinguishes our work from many related works in horizontal FL\cite{fedrec_yang,liang2021fedrec++,lin2020meta,muhammad2020fedfast}. Thus, we focus on discussing most relevant VFL works with similar purposes to ours. FedMVT\cite{kang2020fedmvtsv} is the pioneering work of exploring semi-supervised learning for VFL. It proposed to use the non-overlapping data and validated its effectiveness on multi-view image datasets, while our method explores a different scenario of leveraging the unlabeled overlapping samples and tailored for tabular data. Excluding both the HFL and VFL setting, FedCT\cite{liu2021fedct} firstly studies the cross-domain recommendation problem in the federated transfer learning setting\cite{yang2019federated}. 
However, it is neither suitable for the pure VFL scenario nor capable of leveraging the unlabeled federated data. Besides, these works do not consider the high cost of cross-party inference, which is crucial in advertising. Additionally, another line of work has also put efforts into improving the privacy and security of federated learning\cite{aono2017privacy,liu2021defending,sun2021vertical,jin2021catastrophic}, especially the problem of label leakage\cite{blia_vfl,lia_vfl,lia_protect_vfl,dp_lia}. These ideas are complementary to ours and can be combined to enhance security, while we focus on the aspect of model training in this paper.

\noindent\textbf{Unsupervised Pre-training}.
Self-supervised learning has achieved remarkable success in natural language processing and computer vision ~\cite{devlin2018bert,liu2019roberta,clark2020electra} and is also making progress in recommendation (such as the Correlated Feature Masking
 technique \cite{yao2021self}), but they are all limited to the context of centralized machine learning. Some technics like contrastive learning were also adopted in federated learning \cite{dong2021federated}, but only for the HFL and image data. It is still an open challenge to use extensive unlabeled tabular data for VFL\cite{vfl_survey1,fl_big_survey}.

\noindent\textbf{Knowledge Distillation}. Knowledge distillation \cite{hinton2014distilling} is widely applied in HFL \cite{li2019fedmd,he2020group,lin2020ensemble} to tackle the Non-IID problem. These works require the input fields of the student and teacher must be identical, while in our framework, only part of the input field of the teacher is accessible to a student. This fashion is more related to privileged learning \cite{pri_kd1,pri_kd2,pri_kd3,pri_fed}. However, we further consider the problem with unsupervised pre-training and tailor our method for VFL tasks in particular. Overall, our final student model can leverage information from both labeled and unlabeled data and is capable of making predictions independently.


\section{Method}
\subsection{Preliminary: Two-Party VFL}
We focus on a typical two-party vertical federated learning setting\cite{vepakomma2018split,splitNNVP} where an \textit{\textbf{active party} A} holding the label and some attributes collaborates with a \textit{\textbf{passive party} B} who provides additional attributes to train a federated model. A federated model consists of the \textit{\textbf{bottom model}} holds by each party and the \textit{\textbf{top model}} holds by the active party: 
    \begin{equation}
        \hat{y}_{fed} = g_A(f_A(\mathbf{x}_A), f_B(\mathbf{x}_B))
    \end{equation}
where $f$ denotes the bottom model and $g$ denotes the top model, $\mathbf{x}$ denotes inputs from parties. In the forward propagation, the hidden output $\mathbf{h}_B = f_B(\mathbf{x}_B)$ of party B is sent to party A for the subsequent computation of $g_A$ and loss function; For the backward propagation, the gradient of loss over $\mathbf{h}_B$ is sent to party \textit{B} for the subsequent gradient computation and parameter updates for $f_B$. The transmission process can be combined with encryption techniques to enhance security, as we mentioned in section \ref{sec:rl}.
    
\subsection{Matched Pair Detection}
\subsubsection{\textbf{Task Description}} The matched pair detection (MPD) task aims to learn a binary classifier capable of distinguishing whether the input attributes from two parties are matched. Specifically, we treat all original sample as the positives sample and construct the negatives by in-batch dynamic sampling. Let's use $\mathbf{X} \in \mathcal{R}^{m\times d}$ to denote a batch of inputs from a single party, then a negative batch is constructed by random row permutation :
\begin{align}
    &\mathbf{U_+} = [\mathbf{X}_A; \mathbf{X}_B], \mathbf{U^i_+}=(\mathbf{x}_A^i, \mathbf{x}_B^i)\\
    &\mathbf{U^-} = [\mathbf{P}\mathbf{X}_A; \mathbf{X}_B], \mathbf{P}\in\mathcal{R}^{m\times m}
\end{align}
where $\mathbf{U_+},\mathbf{U_-}$ respectively denotes a positive and negative batch and $\mathbf{P}$ is a \textit{row-permutation matrix} \cite{permu_matrix}. We generate the random matrix $\mathbf{P}$ via frequency-based in-batch random sampling, as a coarse approximation of its global distribution in the whole dataset \cite{neg_pmi}.
Such perturbation can be independently conducted by the active party via matrix multiplication and \textit{\textbf{do not need any cooperation from the passive party}}, which is very friendly for industry federated application. The model is finally trained with both the positive and negative batches: 
\begin{equation}
    \mathcal{L}^{MPD} = \vec{\mathbf{1}}\cdot\left[ \log\sigma(g(\mathbf{U}^+)) + \log\sigma(-g(\mathbf{U}^-)) \right]^\top
\end{equation}
where $g$ is short for the complete federated model. Once the pre-training is finished, only the bottom models $f_A$ and $f_B$ are used as initialization in fine-tuning or distillation.
   
\subsubsection{\textbf{Theoretical Analysis}} In this part, We show MPD implies a intrinsic connection to \textit{\textbf{maximizing mutual information}}. Considering the MPD task is trying to maximize $P(y_u = 1 | \mathbf{x}_A, \mathbf{x}_B)$ for correct pairs while maximizing $P(y_u = 0 | \mathbf{x}_A^*, \mathbf{x}_B)$ for randomly sampled ``negative'' samples. We can re-write the learning objective for a distinct matched pair $(\mathbf{h}_A^i, \mathbf{h}_B^i)$ as:
\begin{equation}\label{eq:neg_obj}
    \mathcal{L}_i = \log\sigma(g_A(\mathbf{h}_A^i, \mathbf{h}_B^i)) + k\cdot\mathbb{E}_{\mathbf{h}_A^*\sim P_D}\log\sigma(-g_A(\mathbf{h}_A^*, \mathbf{h}_B^i))
\end{equation}
   where $\sigma(\cdot)$ is the sigmoid function and $k$ is the number of negative samples, $\mathbf{h}_A^*=f_A(\mathbf{x}_A^*)$ is the hidden representation of a sampled negative half $\mathbf{x}_A^*$, drawn from the empirical ``unigram'' distribution $P_D(\mathbf{h}_A) = P_D(\mathbf{x}_A) = \frac{\#(\mathbf{x}_A)}{|D|}$. Here $\#(\mathbf{x}_A)$ denotes the number of times $\mathbf{x}_A$ appears in the dataset $D$. Similar to word2vec \cite{mikolov2013distributed}, if we analogously treat $\mathbf{h}_A$ and $\mathbf{h}_B$ as a word embedding $\vec w$ and its context embedding $\vec c$, we can find that our equation (\ref{eq:neg_obj}) is isomorphic to equation (1) of \cite{neg_pmi}. The only difference is the choice of similarity function where we use $g_A$ and they use dot-product. Thus we get a similar revelation as in \cite{neg_pmi}:
\begin{equation}\label{eq:max_pmi}
    g_A(\mathbf{h}_A, \mathbf{h}_B) = PMI(\mathbf{x}_A, \mathbf{x}_B) - \log k.
\end{equation}
 That is to say, the top model implicitly models the point-wise mutual information (PMI) of the observed input pairs, with a shifted constant $\log k$. Specifically, when $k=1$, MPD exactly encourages maximizes the mutual information between the hidden representations of two parties. Such learning principle of maximizing (shifted) mutual information strongly supports the MPD pre-training task to learn good cross-party representations. Due to the page limit, we omit the proof and suggest readers refer to \cite{neg_pmi} for more detail.

\label{sec:dataset}    
\subsection{Split Knowledge Distillation}
    To enable independent serving at local, we develop Split Knowledge Distillation (SplitKD) to transfer knowledge from the federated teacher model to a local student model. The student model is aware of passive-party's knowledge but do not rely on its inputs anymore in inference. Let's use superscript $T$ and $S$ to denote teachers and students respectively, the process of distillation can be formalized as:
\begin{align}
    &\hat{y}_{A} = g^S_A(f^S_A(\mathbf{x}_A)) \\
    &\mathcal{L}^{S}_A = \alpha \cdot CE(y, \hat{y}_A) + (1-\alpha)\cdot KL(\hat{y}^T_{Fed}, \hat{y}_A)
\end{align}
where $\hat{y}_{Fed}$, $\hat{y}_A$ denote the predictions, $y$ is the ground-truth label from party A. $g^S_A$ and $f^S_A$ denote the top part and bottom part of the student. $KL(\cdot)$ denotes the KL divergence and $CE(\cdot)$ denotes the binary cross-entropy loss. We use $\alpha$ as a hyper-parameter to balance the effect of raw label and the federated soft label.

Once the distillation is finished, the student model can be independently deployed to existing local inference systems, without any additional system requirements. We emphasize that \textit{\textbf{the deployment of our student model is the same as a traditional local real-time model}}, as long as the current inference system works in real-time (which is already satisfied), we can naturally meet the real-time requirement only by adopting the same model architecture in the student model. 

\begin{figure*}[t]
    \caption{Results on the \textit{\textbf{federated setting}}. It shows that the vanilla VFL outperforms local models in all the datasets and our self-supervised task consistently outperforms all the baselines.``$\uparrow$'' denotes the absolute AUC improvements compared to the baseline. ``U'' denotes the use of unlabeled data. Terms in bold with blue denote the best.}\label{tab:vfl_result}
  \centering
  \begin{minipage}{0.6\textwidth}
  \renewcommand{\arraystretch}{1.15}
    \flushright
    \begin{threeparttable}
        \begin{tabular}{cc|cc|cc|cc}  
        \Xhline{2\arrayrulewidth}
        \textbf{} & \textbf{} & \multicolumn{2}{c|}{{\underline{\textit{\ams}}}} & \multicolumn{2}{c|}{{\underline{\textit{\ola}}}} & \multicolumn{2}{c}{{\underline{\textit{\criteo}}}} \\
        Method & U. & AUC$_\%$ & \myuparrow & AUC$_\%$ & \myuparrow & AUC$_\%$ & \myuparrow \\
        \Xhline{2\arrayrulewidth}
        Baseline Local & \myXm & 70.78 & — & 70.64 & — & 68.30 & — \\
        VFL & \myXm & 72.12 & 1.34 & 72.98 & 2.34 & 75.94 & 7.65 \\
        VFL-SST & \myCm & 72.32 & 1.55 & 73.11 & 2.48 & 76.71 & 8.41 \\
        \textit{\textbf{Ours} (VFL-MPD)} & \myCm & \myColor{\textbf{73.30}} & \myColor{\textbf{2.53}} & \myColor{\textbf{73.31}} & \myColor{\textbf{2.67}} & \myColor{\textbf{77.08}} & \myColor{\textbf{8.79}} \\
        \Xhline{2\arrayrulewidth}
        \end{tabular}
        \begin{tablenotes}
        	\scriptsize
        	\item $\%$: values are shown in percentage for readability.
        \end{tablenotes}
    \end{threeparttable}
    \label{tab:grades}
  \end{minipage}
  \begin{minipage}{0.3\textwidth}
    \flushleft
    \includegraphics[width=4.2cm]{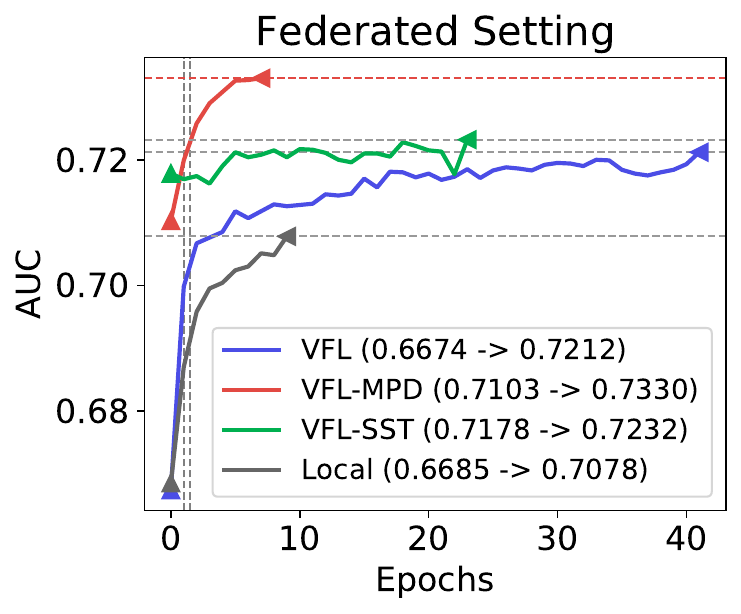}
  \end{minipage}%
\end{figure*}

\begin{figure*}[t]
\caption{Results on the \textit{\textbf{local setting}}. Our method achieves the best result and its two simplified variants also significantly outperform the baseline, indicating the validity of its two components.}\label{tab:vkd_result}
  \centering
  \begin{minipage}{0.6\textwidth}
  \renewcommand{\arraystretch}{1.15}
    \flushright
    \begin{tabular}{cc|cc|cc|cc}
        \Xhline{2\arrayrulewidth}
        \textbf{} & \textbf{} & \multicolumn{2}{c|}{\textit{\underline{\ams}}} & \multicolumn{2}{c|}{\textit{\underline{\ola}}} & \multicolumn{2}{c}{\textit{\underline{\criteo}}} \\
        Method & U. & AUC$_\%$ & \myuparrow & AUC$_\%$ & \myuparrow & AUC$_\%$ & \myuparrow \\
        \Xhline{2\arrayrulewidth}
        Baseline Local & \myXm & 70.78 & — & 70.64 & — & 68.30 & — \\
        Local-SD & \myXm & 71.01 & 0.23 & 71.55 & 0.91 & 68.73 & 0.43 \\
        Local-MPD & \myCm & 70.79 & 0.02 & 71.33 & 0.69 & 68.65 & 0.35 \\
        \textit{\textbf{Ours} (Local-SSD)} & \myCm & \myColor{\textbf{71.67}} & \myColor{\textbf{0.89}} & \myColor{\textbf{71.87}} & \myColor{\textbf{1.23}} & \myColor{\textbf{68.77}} & \myColor{\textbf{0.47}} \\
        \Xhline{2\arrayrulewidth}
    \end{tabular}
  \end{minipage}
  \begin{minipage}{0.3\textwidth}
    \flushleft
    \includegraphics[width=4.1cm]{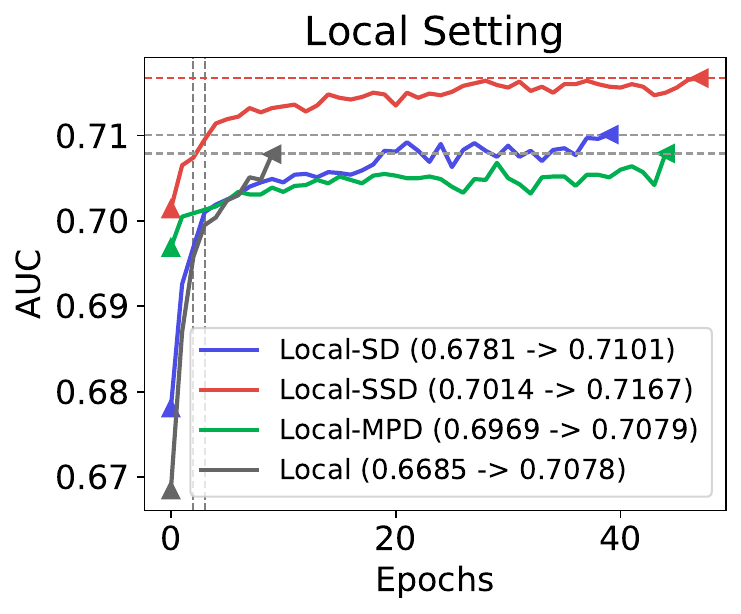}
  \end{minipage}%
\end{figure*}

\section{Experiments}\label{sec:experiment}
We conduct experiments to answer follow questions:
\begin{itemize}[leftmargin=*]
    \item \textit{\textbf{Q1}}: How much does VFL benefits MPD pre-training?
    \item \textit{\textbf{Q2}}: Could student models outperform vanilla local models? 
\end{itemize}
\textbf{Q1} focus on the effectiveness of VFL and MPD in an ideal VFL setting. While \textbf{Q2} focus on benefits of \short in a more practical local serving setting, where federated models are unusable.

\subsection{Experimental Settings}
\subsubsection{\textbf{Datasets}}
We conducted experiments on two industry datasets from Tencent\cite{economist} and a public CTR dataset from Criteo \cite{criteo}. The public dataset is manually partitioned to simulate the federated setting. Both \textit{\textbf{\ams}} and \textit{\textbf{\ola}} are CVR datasets where a game publisher acts as the active party, with a ad platform and a social-network service acts as the passive party correspondingly. The datasets contains user actions on multiple online games (for the active party), various ad items (for \ams) and social activities (in \ola). \textit{\textbf{Criteo-NC}} is acquired by splitting fields of \textit{Criteo} into the \textbf{N}umerical part and the \textbf{C}ategorical part for the active and passive parties, respectively. Details are summarized in Table \ref{tab:dataset}. 



\subsubsection{\textbf{Evaluation Settings}}
We validate our method in two settings. \textit{The federated serving} setting assumes that a real-time federated inference system is available and \textit{the local serving} setting assumes not. Specifically, the active party independently deploy a local model in the local setting but collaboratively deploy the federated model with the passive party in the federated setting.
\begin{itemize}[leftmargin=*]
    \item \textbf{\textit{The federated setting:}} We set 3 methods to validate the advance of leveraging unlabeled data and the efficiency of our pretext task MPD. \textbf{(1) VFL}: The vanilla VFL setting. \textbf{(2) VFL-SST}: We adopt a practice method usually called self-training \cite{st1,st2analyse,st3detection} to leverage federated unlabeled data, as a baseline to compare with our MPD task. Specifically, a federated model pre-trained on the labeled data is used to produce soft labels on the unlabeled data, then a new model is trained with these soft labels. \textbf{(3) VFL-MPD}: We train a federated model on the unlabeled data with the objective of MPD, and then finetune it on the labeled data.
    
    \item \textbf{\textit{The local setting:}} We set 4 methods to validate the effectiveness of the overall framework \short and the performance contributions of MPD and SplitKD: \textbf{(1) Baseline Local}: A simulation scenario without using federated learning where each party trains a local model. The ground truth label is assumed to be shared with the passive party for simulation. \textbf{(2) Local-MPD}: The training of local models is similar to ``Baseline Local'', except that: a) the local bottom model is initialized with its corresponding bottom part in a federated model pre-trained by MPD. b) The bottom part is finetuned. It's a simplified variant of \short to inspect the contribution of MPD. \textbf{(3) Local-SD}: The local models are acquired by splitting up a federated model trained on the labeled data. It's referred to inspect the contribution of SplitKD to \short. \textbf{(4) Local-SSD}: This is the complete version of our method. Compared to SplitKD, the federated teacher is additionally pre-trained on the unlabeled data via MPD.
\end{itemize}

\subsubsection{\textbf{Implementaion Details}} Following related works in advertising \cite{guo2017deepfm}, we use AUC (Area Under the ROC curve) as the evaluation metric. We use a 2-layer MLP for all sub-models in all experiments.
We use the Adam optimizer with $L_2$ regularization for training. The batch size is 10K and 5K for the pre-training task and downstream tasks respectively. We use $1/20$ of training data as the validation set to conduct early stopping. We tune hyper-parameters for all methods and report the best result among 3 repeated runs for each. The main and fine-tune learning rates are chosen in $\eta \in \{1^{-2}, 5*1^{-3}\}, \eta' \in \{1^{-3}, 5*1^{-4}\}$, and the distillation weight $ \alpha \in \{0.5, 0.9\},$ and the $L_2$ penalty coefficient $ \lambda \in \{1^{-4}, 1^{-5}\}$. All experiments are conducted via PowerFL\cite{powerfl}. 

\begin{table}[t]
\renewcommand{\arraystretch}{1.05}
\begin{center}
\caption{Dataset statistics. ``|'' segments values for two parties}\label{tab:dataset}
\begin{tabular}{c|c|c|c}
 \hline
 \textbf{Item} & \textbf{\ams}& \textbf{\ola} & \textbf{\criteo} \\
 \hline
 $\#$\textbf{no label} & 500$\times 10^4$ & 500$\times 10^4$ & 500$\times 10^4$ \\
 $\#$\textbf{labeled} & 64$\times 10^4$ & 36$\times 10^4$ & 15$\times 10^4$ \\
 $\#$\textbf{test} & 2.5$\times 10^4$ & 5$\times 10^4$ & 50$\times 10^4$ \\
 \textbf{positive}$\%$ & 1:13 & 1:20 & 1:4 \\
 \hline
 \hline
 $\#$\textbf{fields} & 51 $|$ 27 & 89 $|$ 39 & 13 $|$ 26 \\
 $\#$\textbf{dim} & 2122 $|$ 1017 & 166 $|$ 298 & 13 $|$ 260 \\
 \hline
\end{tabular}
\end{center}
\end{table}

\subsection{Results and Analysis}

\subsubsection{\textbf{Advantages of Self-Supervised Pre-training}}
As shown in Figure \ref{tab:vfl_result}, the two methods using the unlabeled data get further improvements over vanilla VFL. Besides, our MPD task outperforms ``VFL-SST'' by additional $\mathbf{0.98\%}$, $\mathbf{0.19\%}$, and $\mathbf{0.38\%}$ points on three datasets, respectively. It indicates that MPD can use unlabeled data more efficiently and learns better representation, thus we answered question \textbf{Q1}. Specifically, as additional experiments conducted on \ams dataset shows (right sub-figures of Figure \ref{tab:vfl_result} and Figure \ref{tab:vkd_result}), methods equipped with MPD achieve the best performance improvement curve. 


\subsubsection{\textbf{Advantages of Partial Knowledge Transfer}}
In the setting of local serving, our focal point is the performance of the active party and its improvement achieved by other methods. We evaluate all methods by the absolute AUC improvement over the vanilla local model. As shown in Figure \ref{tab:vkd_result}, our method significantly improves the local performance with \textbf{0.89\%}, \textbf{1.23\%} and \textbf{0.47\%} over three datasets, respectively. The superior performance gives us a sufficient reason to replace the local vanilla model as the student model to conduct online serving. Notice that, the gap between ``Local-SD'' and ``VFL'' is natural, as the set of available features for local models is significantly smaller than for the federated model, as depicted in the middle part of Table \ref{tab:dataset}. We also conducted an ablation study to test the efficacy of the two components of our method. As shown in Figure \ref{tab:vkd_result}, both ''Local-SD'' and ''Local-MPD'' significantly outperform the local baseline, but are weaker than ''Local-SSD''. This suggests that the two components are both efficient. Thus we answered question \textbf{Q2}.

\section{Conclusion and Future Work}
In this paper, we propose a semi-supervised VFL framework VFed-SSD for industry-friendly vertical federated advertising. We highlight the existence of massive unlabeled overlapping data in advertising and develop a self-supervised task MPD to exploit its potential in representation learning. Meanwhile, we develop a novel knowledge transfer schema SplitKD to meet the strict system response time restriction of advertising systems. Experiments on both industry and public benchmark datasets validate the effectiveness of our method. In the future, we plan to extend our work to other tabular-data-based VFL tasks and the multi-party scenarios. 


\bibliographystyle{ACM-Reference-Format}
\bibliography{main.bib}





\end{document}